\documentclass[letterpaper, 10 pt, conference]{ieeeconf}  

\IEEEoverridecommandlockouts                              

\usepackage{graphics} 
\usepackage{graphicx,subfigure}
\usepackage{epsfig} 
\usepackage{amsmath} 
\usepackage{amssymb}  
\usepackage{multirow} 
\usepackage{algorithm}
\usepackage{algorithmic}
\usepackage{stfloats}
\usepackage{booktabs}
\usepackage{xcolor}
\usepackage{hyperref}

\title{\LARGE \bf
Virtual-to-real Deep Reinforcement Learning: \\
Continuous Control of Mobile Robots for Mapless Navigation
}

\author{Lei Tai$^{1,2}$ 
and Giuseppe Paolo$^{3}$
and Ming Liu$^{2}$
\thanks{$^{*}$This paper is supported by Shenzhen Science, Technology and Innovation Commission (SZSTI) JCYJ20160428154842603 and JCYJ20160401100022706; partially supported by the Research Grant Council of Hong Kong SAR Government, China, under Project No. 21202816 and No. 16212815, partially supported by the HKUST Project IGN16EG12; awarded to Prof. Ming Liu.}
\thanks{$^{1}$MBE, City University of Hong Kong; $^{2}$ECE, the HKUST; $^{3}$D-MAVT, ETH Zurich. {\tt\small ltai@ust.hk, eelium@ust.hk, giupaolo@student.ethz.ch}}
}

\begin{document}

\maketitle
\thispagestyle{empty}
\pagestyle{empty}

\begin{abstract}
We present a learning-based mapless motion planner by taking the sparse 10-dimensional range findings and the target position with respect to the mobile robot coordinate frame as input and the continuous steering commands as output. Traditional motion planners for mobile ground robots with a laser range sensor mostly depend on the obstacle map of the navigation environment where both the highly precise laser sensor and the obstacle map building work of the environment are indispensable. We show that, through an asynchronous deep reinforcement learning method, a mapless motion planner can be trained end-to-end without any manually designed features and prior demonstrations. The trained planner can be directly applied in unseen virtual and real environments. The experiments show that the proposed mapless motion planner can navigate the nonholonomic mobile robot to the desired targets without colliding with any obstacles. 
\end{abstract}

\section{Introduction}
%

\subsubsection{Deep Reinforcement Learning in mobile robots}
Deep Reinforcement Learning (deep-RL) methods achieve great success in many tasks including video games \cite{mnih2015human} and simulation control agents \cite{lillicrap2015continuous}. The applications of deep reinforcement learning in robotics are mostly limited in manipulation \cite{gu2016continuous} where the workspace is fully observable and stable. In terms of mobile robots, the complicated environments enlarge the sample space extremely while deep-RL methods normally sample the action from a discrete space to simplify the problem \cite{zhu2016target, sadeghi2016cad, tai2016towards}. Thus, in this paper, we focus on the navigation problem of nonholonomic mobile robots with continuous control of deep-RL, which is the essential ability for the most widely used robot.
 
   \begin{figure}[!tp]
      \centering
      \includegraphics[width=\columnwidth]{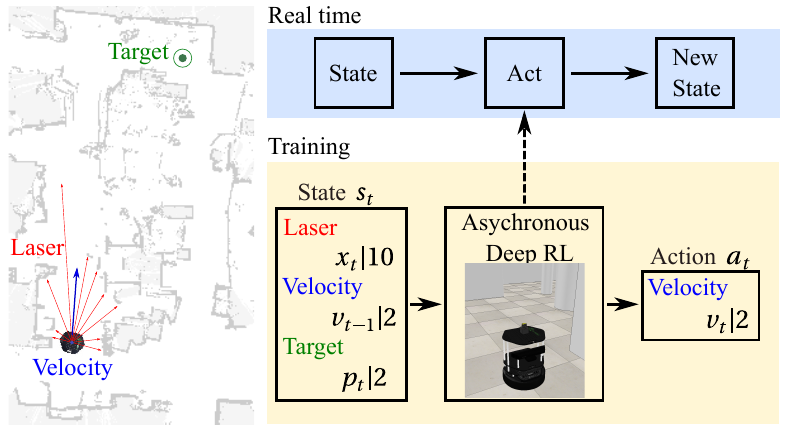}
      \caption{A mapless motion planner was trained through asynchronous deep-RL to navigate a nonholonomic mobile robot to the target position collision free. The planner was trained in the virtual environment based on sparse 10-dimensional range findings, 2-dimensional previous velocity, and 2-dimensional relative target position. 
      }
      \label{fig:no_1}
      \vspace{-2em}
   \end{figure}

\subsubsection{Mapless navigation}
Motion planning
aims at navigating robots to the desired target from the current position without colliding with obstacles. For mobile nonholonomic ground robots, traditional methods, like simultaneous localization and mapping (SLAM), handle this problem through the prior obstacle map of the navigation environment \cite{durrant2006simultaneous} based on dense laser range findings. Manually designed features are extracted to localize the robot and build the obstacle map. There are two less addressed issues for this task: (1) the time-consuming building and updating of the obstacle map, and (2) the high dependence on the precise dense laser sensor for the mapping work and the local cost-map prediction. It is still a challenge to rapidly generate appropriate navigation behaviors for mobile robots without an obstacle map and based on sparse range information.

Nowadays, low-cost methods, like  WiFi localization \cite{sun2014wifi} and visible-light communication \cite{qiu2016let}, provide lightweight solutions for mobile robot localization. Thus, mobile robots are able to get the real-time target position with respect to the robot coordinate frame. And it is really challenging for a motion planner to generate global navigation behaviors with the local observation and the target position information directly without a global obstacle map.
Thus, we present a learning-based mapless motion planner. In virtual environments, a nonholonomic differential drive robot was trained to learn how to arrive at the target position with obstacle avoidance through asynchronous deep reinforcement learning \cite{mnih2016asynchronous}. 

\subsubsection{From virtual to real world}
Most of the training of deep-RL is implemented in a virtual environment because the trial-and-error training process may lead to unexpected damage to the real robot for specific tasks, like obstacle avoidance in our case. The huge difference between the structural simulation environment and the highly complicated real-world environment is the central challenge to transfer the trained model to a real robot directly. In this paper, we only used 10-dimensional sparse range findings as the observation input. This highly abstracted observation was sampled from specific angles of the raw laser range findings based on a trivial distribution. This brings two advantages: the first is the reduction of the gap between the virtual and real environments based on this abstracted observation, and the second is the potential extension to low-cost range sensors with distance information from only 10 directions. 









We list the main contributions of this paper:
(1) A mapless motion planner was proposed by taking only 10-dimensional sparse range findings and target relative information as references. (2) The motion planner was trained end-to-end from scratch through an asynchronous deep-RL method. The planner can output continuous linear and angular velocities directly. (3) The learned planner can generalize to a real nonholonomic differential robot platform without any fine-tuning to real-world samples.


 
\section{Related Work} \label{sec:rel}
\subsection{Deep-Learning-based navigation}
Benefiting from the improvement of high-performance computational hardware, deep neural networks show great potential for solving complex estimation problems. For learning-based obstacle avoidance, deep neural networks have been successfully applied on monocular images \cite{lecun2005off} and depth images \cite{tai2016deep}. Chen \textit{et al.} \cite{chen2015deepdriving} used semantics information extracted from the image by deep neural networks to decide the behavior of the autonomous vehicle. However, their control commands are simply discrete actions like \textit{turn left} and \textit{turn right} which may lead to rough navigation behaviors.

Regarding learning from demonstrations, Pfeiffer \textit{et al.} \cite{pfeiffer2016perception} used a deep learning model to map the laser range findings and the target position to the moving commands. 
Kretzschmar \textit{et al.} \cite{kretzschmar2016socially} used inverse reinforcement learning methods to make robots interact with humans in a socially compliant way. Such kinds of trained models are highly dependent on the demonstration information. A time-consuming data collection procedure is also inevitable. 
\subsection{Deep Reinforcement Learning}
Reinforcement learning has been widely applied in robotic tasks \cite{kober2013reinforcement, tai2016deep_survey}. Minh \textit{et al.} \cite{mnih2015human} utilized deep neural networks for the function estimation of value-based reinforcement learning which was called deep Q-network (DQN). 
Zhang \textit{et al.} \cite{zhang2016deep} provided a solution for robot navigation based on depth image trained with DQN, where successor features were used to transfer the strategy to unknown environment efficiently.
The original DQN can only be used in tasks with a discrete action space. To extend it to continuous control, Lillicrap \textit{et al.} \cite{lillicrap2015continuous} proposed deep deterministic policy gradients (DDPG) to use deep neural networks on the actor-critic reinforcement learning method where both the policy and value of the reinforcement learning were represented through hierarchical networks. Gu \textit{et al.} \cite{gu2016continuous} proposed continuous DQN based on the normalized advantage function (NAF). The successes of these deep-RL methods are mainly attributed to the \textit{memory replay} strategy in fact. As off-policy reinforcement learning methods, all of the transitions can be used repeatedly. Therefore, asynchronous deep-RL with multiple sample collection threads working in parallel should improve the training efficiency of the specific policy significantly. Gu \textit{et al.} \cite{gu2016deep} proposed asynchronous NAF and trained the model with real-world samples where a door opening task was accomplished by a real robot arm.

A less addressed issue for off-policy methods is the enormous requirement for data sampling. Minh \textit{et al.} {\cite{mnih2016asynchronous} optimize the deep-RL with asynchronous gradient descent from parallel on-policy actor-learners (A3C). Based on this state-of-the-art deep reinforcement learning method,
Mirowski \textit{et al.} \cite{mirowski2016learning} trained a simulated agent to learn navigation in a virtual environment through raw images. Loop closure and depth estimation were proposed as well through parallel supervised learning, but the holonomic motion behavior was difficult to transfer to the real environment. Zhu \textit{et al.} \cite{zhu2016target} trained an image-based planner where the robot learned to navigate to the referenced image place based on the instant view. However, they defined a discrete action space to simplify the task. On the other hand, A3C needs several parallel simulation environments, which limited its extension to some specific simulation engine like \textit{V-REP} \cite{rohmer2013v} which can not be paralleled in the same machine. Thus, we choose DDPG as our training algorithm. Compared with NAF, DDPG needs less training parameters. And we extend DDPG to an asynchronous version as \cite{gu2016deep} to improve the sampling efficiency.

Generally, this paper focuses on developing a mapless motion planner based on low-dimensional range findings. We believe that this is the first time a deep-RL method being applied on the real world continuous control of differential drive mobile robots for navigation. 

\section{Motion planner implementation} \label{sec:app}
\subsection{Asynchronous Deep Reinforcement Learning}
\label{sec:ADDPG}

Compared with the original DDPG, we separate the sample collecting process to another thread as in \cite{gu2016deep}, called Asynchronous DDPG (ADDPG). It can also be implemented with multiple data collection threads as other asynchronous methods. 

   \begin{figure}[!h]
      \centering
      \includegraphics[width=\columnwidth]{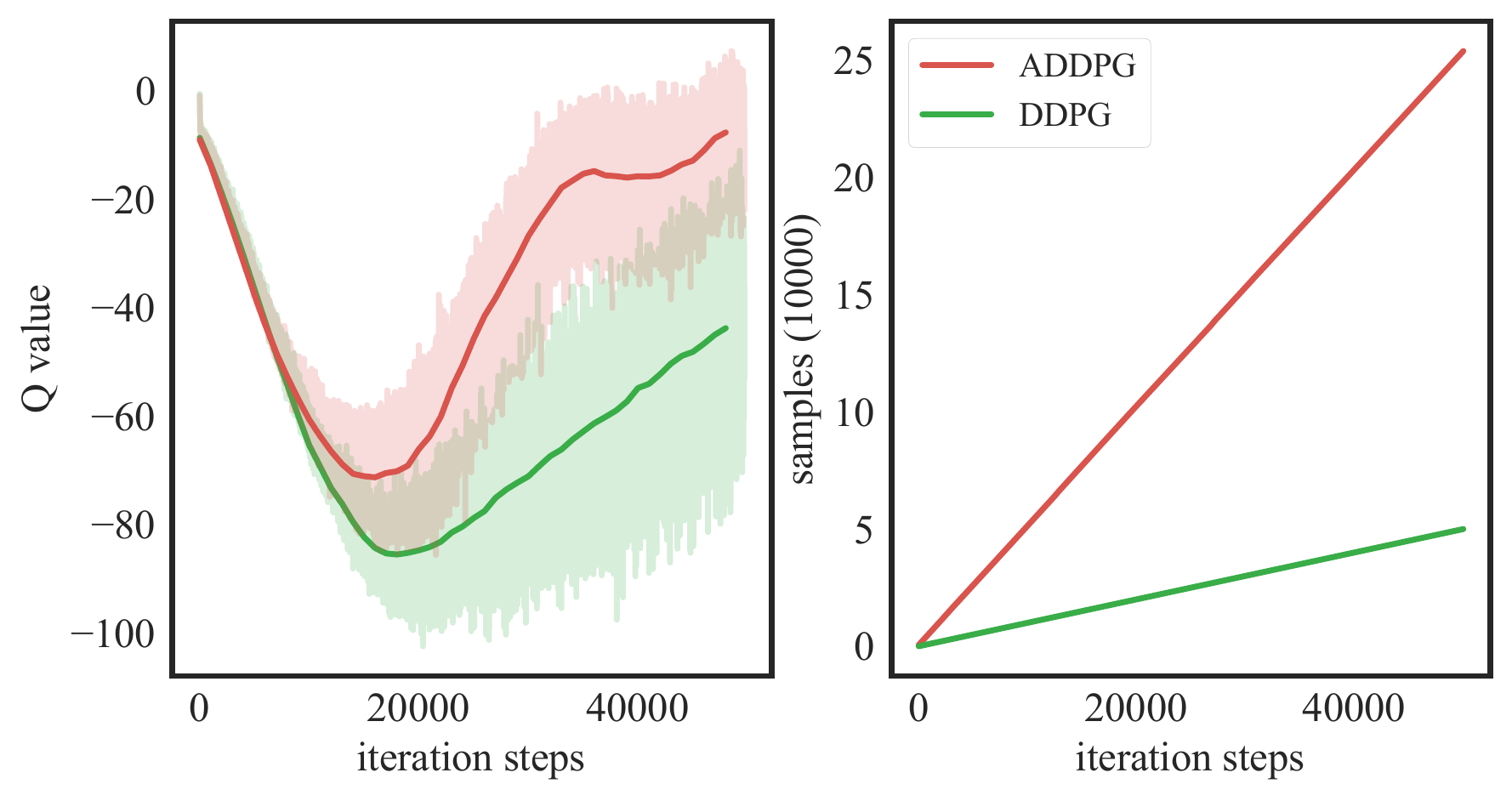}
      \vspace{-2em}
      \caption{Effectiveness test of the Asynchronous DDPG algorithm on the OpenAI Gym task \textit{Pendulum-v0}. Mean Q-value of the training batch in every back-propagation iteration step is shown. The right figure is the count of samples collected with the iteration steps increasing.}
      \label{fig:testaddpg}
      \vspace{-1em}
   \end{figure}
   
To show the effectiveness of the ADDPG algorithm, we tested it in an OpenAI Gym task \textit{Pendulum-v0} 
with one sample collecting thread and one training thread. Trivial neural network structures were applied on the actor and critic networks of this test model. The result is presented in Fig. \ref{fig:testaddpg}. 
The increasing of the Q-value of ADDPG is much faster than the original DDPG, which means ADDPG is able to learn the policy to finish this task in different states more efficiently. This is mainly attributed to the samples collection thread in parallel. As shown on the right of Fig. \ref{fig:testaddpg}, the original DDPG collects one sample every back-propagation iteration while the parallel ADDPG collects almost four times more samples than the original DDPG in every step. 
\subsection{Problem Definition}

This paper aims to provide a mapless motion planner for mobile ground robots. We try to find such a translation function:
\[  v_{t}=f(x_t,p_t,v_{t-1}),\]
where $x_t$ is the observation from the raw sensor information, $p_t$ is the relative position of the target, and $v_{t-1}$ is the velocity of the mobile robot in the last time step. They can be regarded as the instant state $s_t$ of the mobile robot. The model directly maps the state to the action, which is the next time velocity $v_{t}$, as shown in Fig \ref{fig:no_1}. As an effective motion planner, the control frequency must be guaranteed so that the robot can react to new observations immediately. 
\subsection{Network Structure}

The problem can be naturally transferred to a reinforcement learning problem. 
In this paper, we use the extend asynchronous DDPG \cite{lillicrap2015continuous} to train our model as described in Section \ref{sec:ADDPG}. 
   \begin{figure}[!h]
      \centering
      \includegraphics[width=\columnwidth]{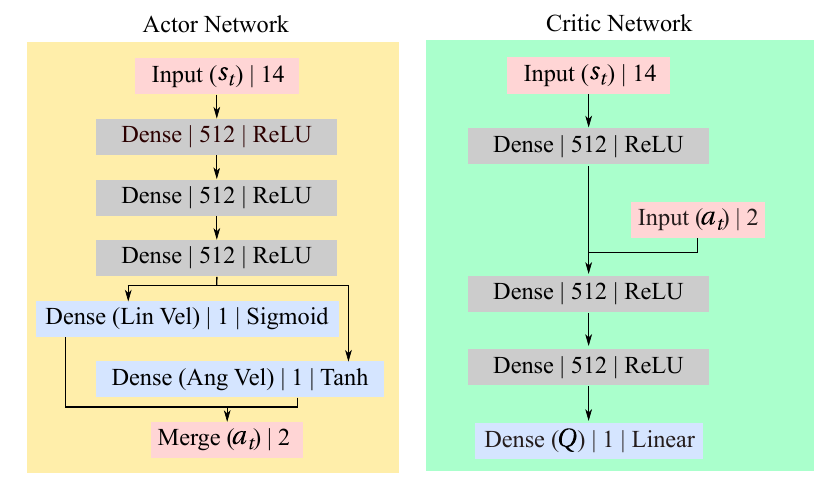}
      \caption{The network structure for the ADDPG model. Every layer is represented by its type, dimension and activation mode. Notice that the \textit{Dense} layer here means a fully-connected neural network. The \textit{Merge} layer simply combines the several input blobs into a single one.}
      \label{fig:network_structure}
      \vspace{-1em}
   \end{figure}

As presented in Fig. \ref{fig:no_1} and the definition function, the abstracted 10-dimensional laser range findings, the previous action, and the relative target position are merged together as a 14-dimensional input vector. The sparse laser range findings are sampled from the raw laser findings between -90 and 90 degrees in a trivial and fixed angle distribution. The range information is normalized to (0,1). The 2-dimensional action of every time step includes the angular and the linear velocities of the differential mobile robot. The 2-dimensional target position is represented in polar coordinates (distance and angle) with respect to the mobile robot coordinate frame. As shown in Fig. \ref{fig:network_structure}, after 3 fully-connected neural network layers with 512 nodes, the input vector is transferred to the linear and angular velocity commands of the mobile robot. To constrain the range of angular velocity in $(-1,1)$, a hyperbolic tangent function (\textit{tanh}) is used as the activation function. Moreover, the range of the linear velocity is constrained in $(0, 1)$ through a \textit{sigmoid} function. Backward moving is not expected because laser findings cannot cover the back area of the mobile robot. The output actions are multiplied with two hyper parameters to decide the final linear and angular velocities directly executed by the mobile robot. Considering the real dynamics of a \textit{Turtlebot}, we choose $0.5\ m/s$ as the max linear velocity and $1\ rad/s$ as the max angular velocity.

For the critic-network, the Q-value of the state and action pair is predicted. We still use 3 fully-connected neural network layers to process the input state. The action is merged in the second fully-connected neural network layers. The Q-value is finally activated through a linear activation function:
\[y = kx+b, \]
where $x$ is the input of the last layer, $y$ is the predicted Q-value, and $k$ and $b$ are the trained weights and bias of this layer. 

\subsection{Reward Function Definition}

There are three different conditions for the reward directly used by the critic network without clipping or normalization:

\[ 
r(s_t, a_t) = 
  \left\{
   \begin{array}{cl}
    \ &r_{arrive} \ \text{if} \ d_t < c_d \\
    \ &r_{collision} \ \text{if} \ min_{x_t} < c_o \\
    \ &c_r (d_{t-1}-d_{t}) \\
    \end{array}
  \right.
\]
If the robot arrives at the target through distance threshold checking, a positive reward $r_{arrive}$ is arranged, but if the robot collides with the obstacle through a minimum range findings checking, a negative reward $r_{collision}$ is arranged. Both of these two conditions stop the training episode. Otherwise, the reward is the difference in the distance from the target compared with last time step, $d_{t-1} - d_{t}$, multiplied by a hyper-parameter $c_r$. This motivates the robot to get closer to the target position. 

\section{Experiments} \label{sec:exp}
\subsection{Training in simulation}
  \begin{figure}[!ht]
      \centering
      \includegraphics[width=\columnwidth]{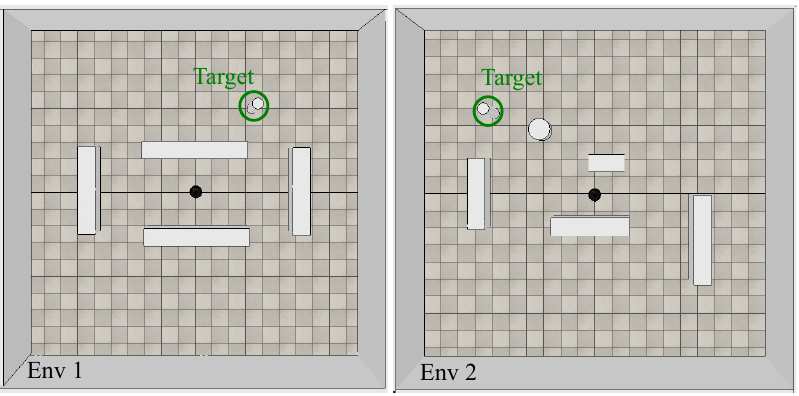}
      \caption{The virtual training environments were simulated by \textit{V-REP} \cite{rohmer2013v}. We built two $10 \times 10 \ m^2$  indoor environments with walls around them. Several different shaped obstacles were located in the environments. A \textit{Turtlebot} was used as the robotics platform. The target labeled in the image is represented by a cylinder object for visual purposes, but it cannot be rendered by the laser sensor. \textit{Env-2} is more compact than \textit{Env-1}.}
      \label{fig:trainenv}
      \vspace{-1em}
   \end{figure}
The training procedure of our model was implemented in virtual environments simulated by \textit{V-REP} \cite{rohmer2013v}. We constructed two indoor environments to show the influence of the training environment on the motion planner, as shown in Fig. \ref{fig:trainenv}. Obstacles in \textit{Env-2} are more compact around the robot initial position. Both models of these two environments were learned from scratch.
A \textit{Turtlebot} is used as the robot platform. The target is represented by a cylinder object, as labeled in the figure. In fact, it cannot be rendered by the laser sensor mounted on the \textit{Turtlebot}. In every episode, the target position was initialized randomly in the whole area and guaranteed to be collision-free with other obstacles. 


The learning rates for the critic and actor network are the same as 0.0001 where the hyper-parameters for the reward function were set trivially. Moreover, the experiments result also show that the effects of ADDPG are not depending on the tuning of hyperparameters. We trained the model from scratch with an Adam \cite{kingma2014adam} optimizer on a single Nvidia GeForce GTX 1080 GPU
for $0.8$m training steps which took almost $20$ hours.
   \begin{figure}[!h]
      \centering
      \includegraphics[width=0.8\columnwidth]{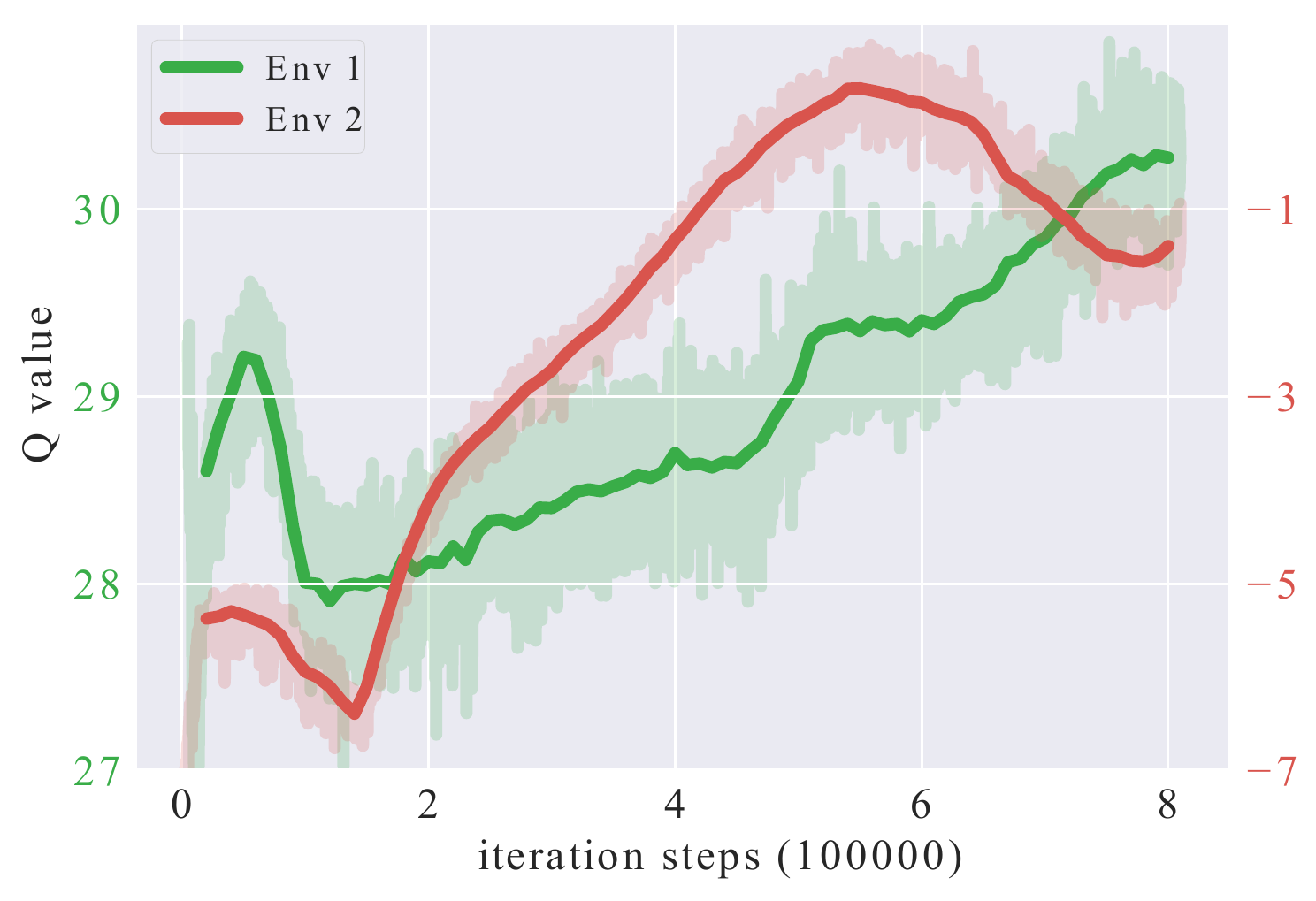}
      \caption{Mean Q-value of the training batch samples in every training step. Notice that two curves from different environments use different y-axes.}
      \vspace{-1em}
      \label{fig:trainingq}
   \end{figure}

The mean Q-value of the training batch samples of the two environments is shown in Fig. \ref{fig:trainingq}. The compact environment \textit{Env-2} received more collision samples, so the Q-value is much smaller than the \textit{Env-1}, but the mean Q-value of \textit{Env-2} increases much faster than \textit{Env-1}.

\subsection{Evaluation}

\begin{figure}[!ht]
  \centering
     \subfigure[Platform]{\includegraphics[width=0.32\columnwidth]{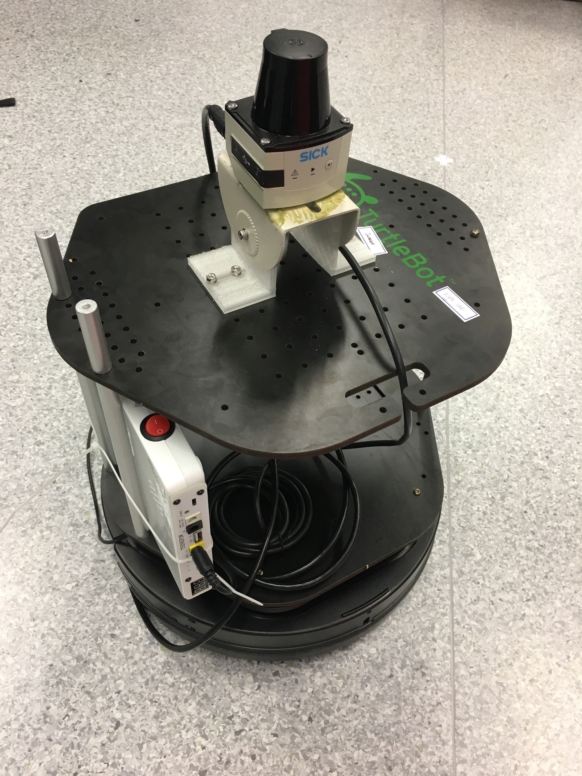}
    \label{fig:platform}}
      \subfigure[Pipeline in real time]{\includegraphics[width=0.5\columnwidth]{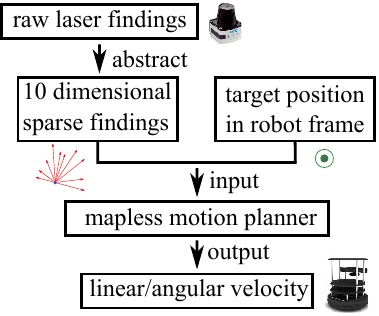}
    \label{fig:realtime_pip}}  
  \caption{The robotics platform is a Kobuki based \textit{Turtlebot}. A SICK TiM570 laser range finder is mounted on the robot. A laptop with an Intel Core i7-4700 CPU is used on-board. Notice that only 10-dimensional sparse range findings extracted from the raw laser findings were used in the real time evaluation as shown in Fig. \ref{fig:realtime_pip} for the baseline planner and deep-RL trained planners. }
  \vspace{-1em}
\end{figure}
\begin{figure*}[!ht]
      \centering
     \subfigure[\textit{Move Base}]{\includegraphics[width=0.49\columnwidth]{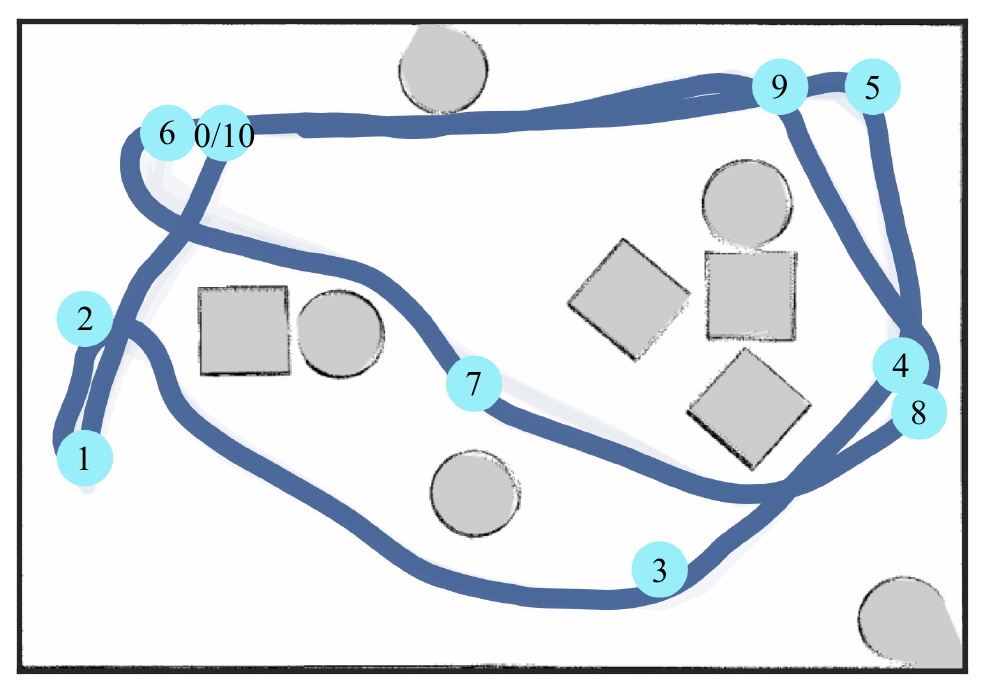}
    \label{fig:movebase_simu}}
      \subfigure[10-dim \textit{Move Base}]{\includegraphics[width=0.49\columnwidth]{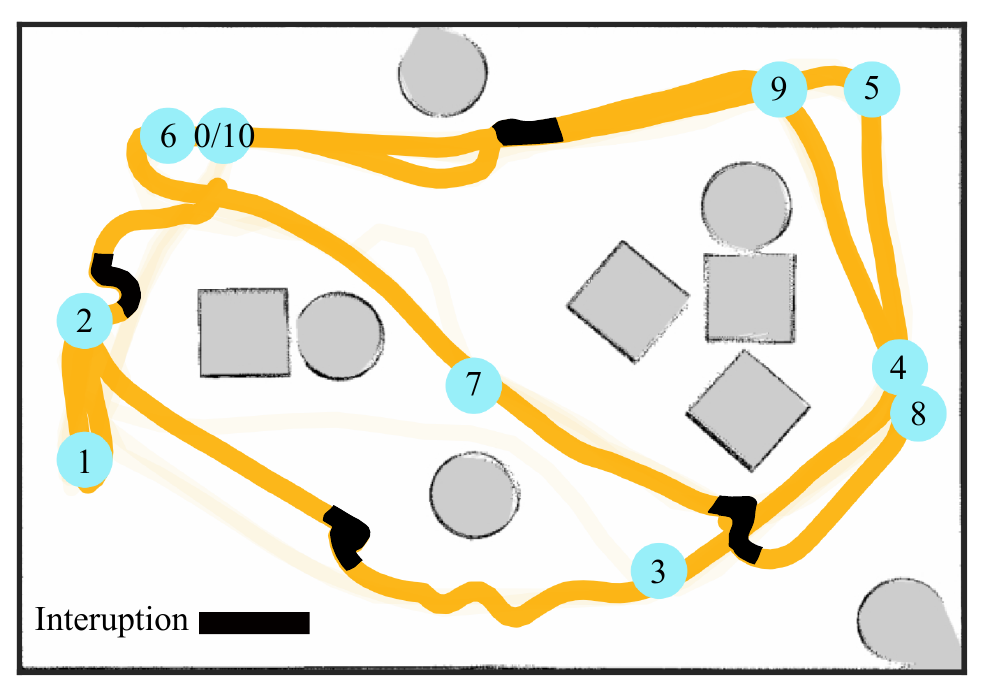}
    \label{fig:noise}}   
      \subfigure[\textit{Env-1}]{\includegraphics[width=0.49\columnwidth]{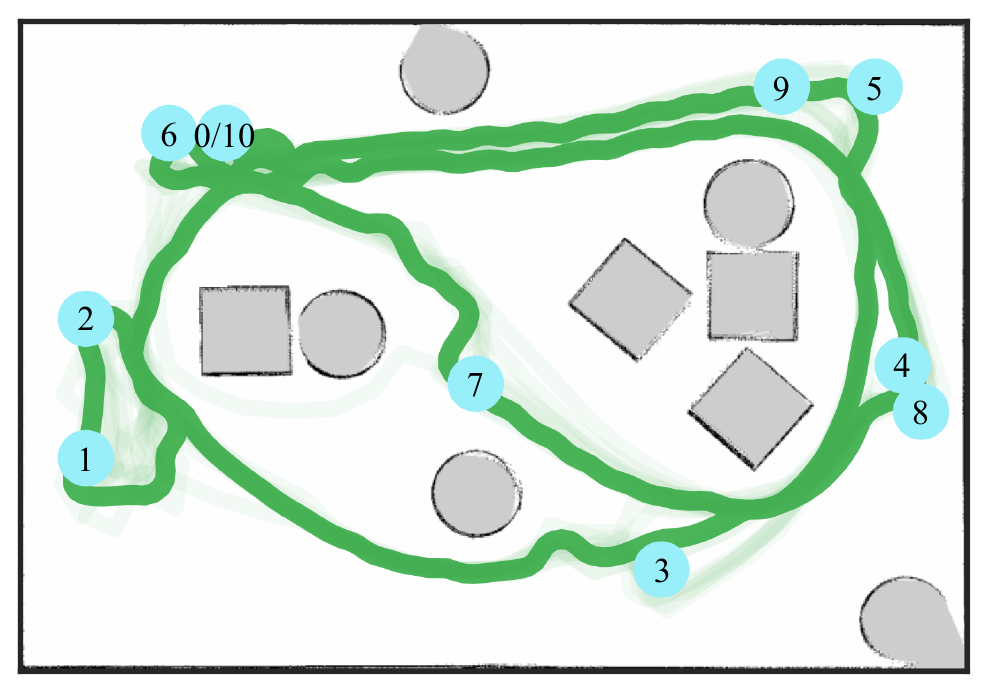}
    \label{fig:policy1}}   
     \subfigure[\textit{Env-2}]{\includegraphics[width=0.49\columnwidth]{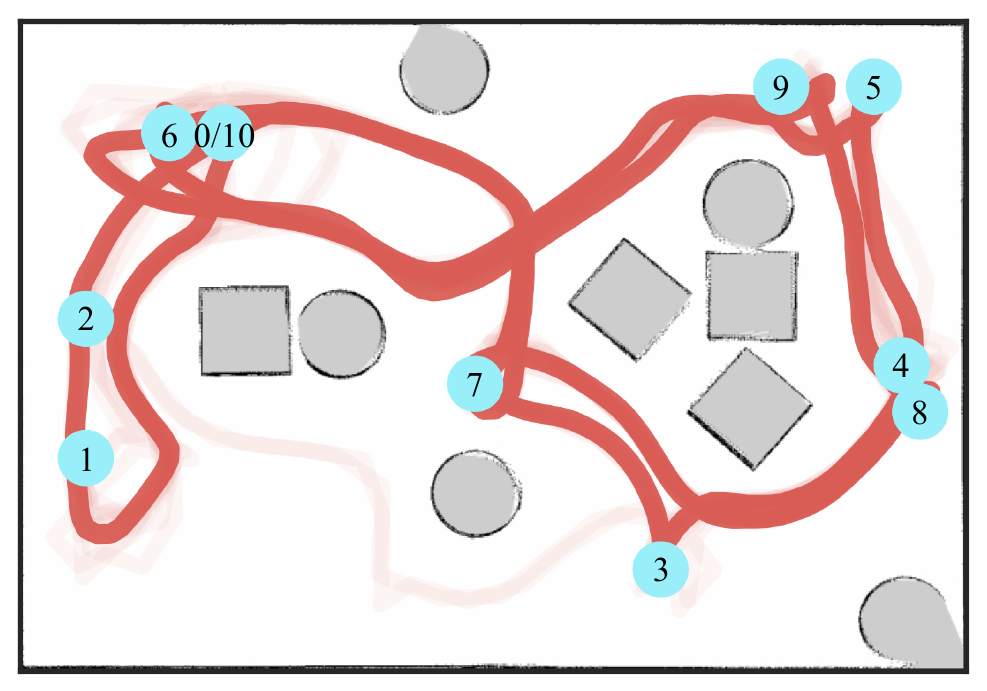}
    \label{fig:policy2}}  
      \caption{Trajectory tracking in the virtual test environment. (a) Original \textit{Move Base}, (b) 10-dimensional \textit{Move Base}, and deep-RL trained models in (c) \textit{Env-1} and (d) \textit{Env-2} are compared. 10-dimensional \textit{Move Base} was not able to finish the navigation tasks. 
      }
      \label{fig:simutest}
\end{figure*}

\begin{figure*}[!ht]
  \centering
  \includegraphics[width=1.5\columnwidth]{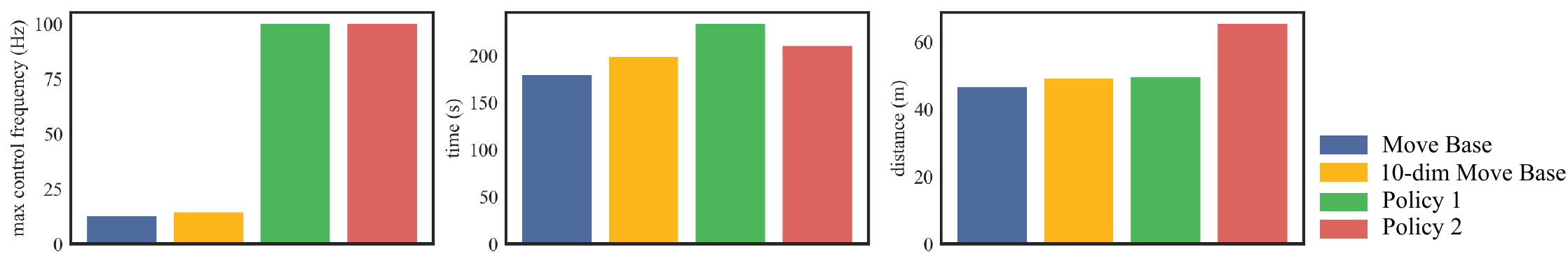}
  \caption{Quantity evaluations between the baseline motion planner and the proposed mapless motion planner, including max control frequency, traveling time, and traveling distance.}
  \label{fig:simutestquan}
  \vspace{-2em}
\end{figure*}

To show the performance of the motion planner when it is deployed on a real robot, we tested it both in the virtual and real worlds on a laptop with an Intel Core i7-4700 CPU. We used a Kobuki based \textit{Turtlebot} as the mobile ground platform.    
The robot subscribed the laser range findings from a SICK TiM551
which has a field of view (FOV) of $270^{\circ}$ and an angular resolution of $0.33^{\circ}$. The scanning range is from $0.05m$ to $10m$, when implemented in the real world, as shown in Fig. \ref{fig:platform}. This paper mainly introduces the mapless motion planner so we did not test the planner effects with different localization methods. The real time position of the robot was provided by \textit{amcl} 
to calculate the polar coordinates of the target position. 

\subsubsection{Baseline}
We compared the deep-RL trained motion planners with the state-of-art \textit{Move Base} motion planner. \textit{Move Base} uses the full laser range information for local cost-map calculation, while our mapless motion planner only needs 10-dimensional sparse range findings from specific directions for motion planning. Therefore, we implemented a 10-dimensional \textit{Move Base} using the laser range findings from specific angles as the trained model, as shown in Fig. \ref{fig:realtime_pip}. These 10-dimensional findings were extended to an 810-dimensional vector covering the field of view through an RBF kernel Gaussian process regression \cite{rasmussen2006gaussian} for the local cost-map prediction that we called 10-dimensional \textit{Move Base} in the following experiments. 
Here the deep-RL trained models only considered the final position of the robot but not the orientation of the desired target.

\subsubsection{Virtual Environment Evaluation}
To show the generic adaptability of the model in other environments, we first built a virtual test environment, as shown in Fig. \ref{fig:simutest}, consisting of a $7\times10m^2$ area with multiple obstacles. We set 10 target positions for the motion planner. The motion planner should navigate the robot to the target positions along the sequence number. For \textit{Move Base}, an obstacle map of the global environment should be built before the navigation task so that the planner can calculate the path. For our deep-RL trained models, as shown in Fig \ref{fig:policy1} and Fig \ref{fig:policy2}, the map is only for trajectory visualization. 

The trajectory tracking of the four planners is shown in Fig. \ref{fig:simutest} as a qualitative evaluation. Every motion planner was executed five times for all of the target positions from $0$ to $10$ in order and one of the paths is highlighted in the figure.

As shown in \ref{fig:noise}, the 10-dimensional \textit{Move Base} cannot finish the navigation task successfully: the navigation was interrupted and aborted because of incorrect prediction of the local cost-map so the robot was not able to find the path by itself and human intervention had to be added to help the robot finish all of the navigation tasks. The interruption parts are labeled as black segments in \ref{fig:noise}. However, deep-RL trained mapless motion planners accomplished the tasks collision free, as shown in Fig. \ref{fig:policy1} and Fig. \ref{fig:policy2}. The deep-RL trained planners show great adaptability to unseen environments. We chose three metrics to evaluate the different planners quantitatively, as listed in Fig. \ref{fig:simutestquan}: (1) \emph{max control frequency}: max moving commands output times per minute. (2) \emph{time}: traveling time for all of the 10 target positions. (3) \emph{distance}: path distance for all of the 10 target positions.

The max control frequency reflects the query efficiency of the motion planner: the query of trained mapless motion planners only took almost $1ms$ which is 7 times faster than the map-based motion planner. Compared with the 10-dimensional 
\textit{Move Base}, \textit{Env-2} took almost the same time to finish all of the navigation tasks even though the path was not the optimally shortest. The \textit{Env-1} motion planning results seem not as smooth as the other motion planners.


\subsubsection{Real Environment Evaluation}
We implemented a similar navigation task but in the real world environment. According to the trajectory in Fig. \ref{fig:simutest}, the motion planner trained in \textit{Env-2} generated a smoother trajectory than the one trained in \textit{Env-1}, and the \textit{Env-2} model seemed to be more sensitive to the obstacles than the \textit{Env-1} model. Preliminary experiments in the real world showed that the \textit{Env-1} model was not able to finish the navigation task successfully. So we only compared the trajectory in the real world between 10-dimensional \textit{Move Base} and the \textit{Env-2} model. We navigated the robot in a complex indoor office environment, as shown in Fig. \ref{fig:realtest}. The robot should arrive at the targets based on the sequence from $0$ to $9$ labeled in the figure.
   \begin{figure*}[!ht]
      \centering
           \subfigure[10-dim \textit{Move Base}]{\includegraphics[width=0.9\columnwidth]{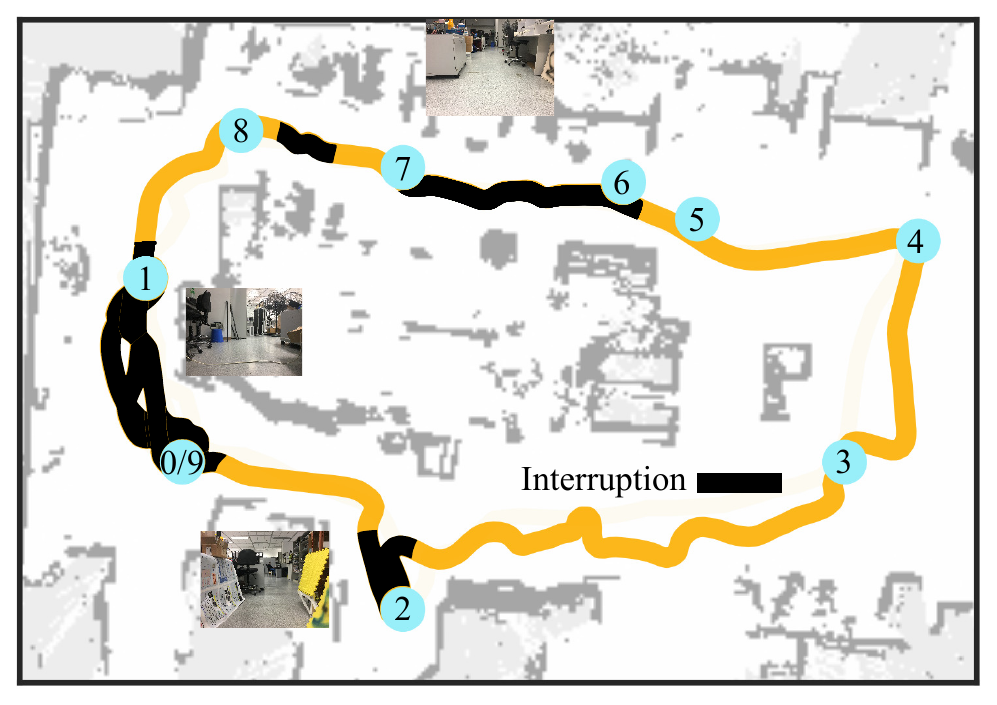}
    \label{fig:movebase_real}}
         \subfigure[\textit{Env-2}]{\includegraphics[width=0.9\columnwidth]{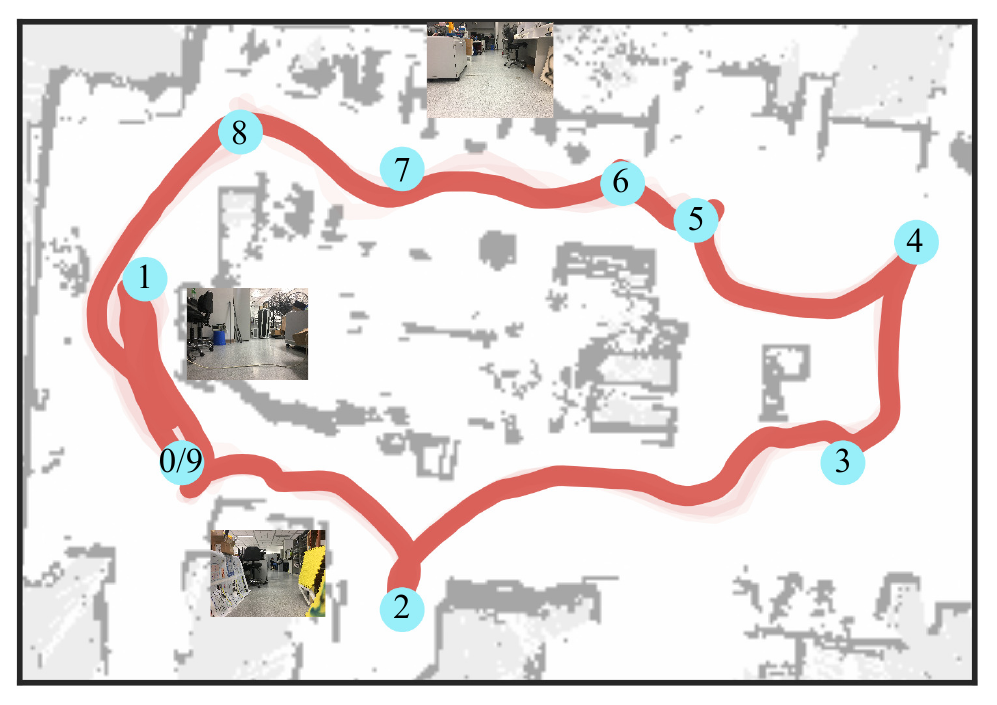}
    \label{fig:policy_2_real}}
          \caption{Trajectory tracking in the real test environment. 10-dimensional \textit{Move Base}, and the deep-RL trained model in \textit{Env-2} are compared. 10-dimensional \textit{Move Base} was not able to finish the navigation tasks. Human innervations were added labled as black segments in Fig. \ref{fig:movebase_real}. }
      \label{fig:realtest}
      \vspace{-2em}
   \end{figure*}

From the trajectory figure, 10-dimensional \textit{Move Base} cannot go across the seriously narrow area because of the misprediction of the local cost-map based on the limited range findings. 10-dimensional \textit{Move Base} was not able to find an effective path to arrive at the desired target. We added human intervention to help the 10-dimensional \textit{Move Base} finish the navigation task. The intervention segments of the path are labeled in black in Fig \ref{fig:realtest}.

The \textit{Env-2} model was able to accomplish all of the tasks successfully. However, sometimes the robot was not able to go across the narrow route smoothly. A \textit{recovery} behavior like the rotating recovery in \textit{Move Base} was developed by the mapless motion planner. Even then, obstacle collision never happened for the mapless motion planner. A brief video about the performance of the mapless planner in different training stages and in test environments is available at \url{https://youtu.be/9AOIwBYIBbs}.

\section{Discussion} \label{sec:dis}

The experiments in the virtual and real world proved that the deep-RL trained mapless motion planner can be transferred directly to unseen environments. The different navigation trajectories of the two training environments showed that the trained planner is influenced by the environment to some extent. \textit{Env-2} is much more aggressive with closer obstacles so that the robot can navigate in the complex real environment successfully. 

In this paper, we compared our deep learning trained model with the original and low-dimensional map-based motion planner. 
Compared with the trajectories of \textit{Move Base}, the path generated from our planner is more tortuous. A possible explanation is that the network has neither the memory of the previous observation nor the long-term prediction ability. Thus LSTM and RNN \cite{hausknecht2015deep} are possible solutions for that problem. We set this revision as future work.

However, we are not aiming to replace the map-based motion planner: it is obvious when the application scenario is a large-scale and complex environment, the map of the environment can always provide a reliable navigation path. Our target is to provide a low-cost solution for an indoor service robot with several range sensors, like a light-weight sweeping robot. The experiments showed that \textit{Move Base} with sparse range findings can not be adapted to narrow indoor environments. Although we used the range findings from a laser sensor, it is certain that this 10-dimensional information can be replaced by low-cost range sensors.

In addition, reinforcement learning methods provide a considerable online learning strategy. The effects of the motion planner can be developed considerably with training in different environments continuously. In this developing procedure, no feature revision or human labeling is needed. On the other hand, the application of the deep neural networks provides a solution for multiple sensor inputs like RGB image and depth. The proposed model has shown the ability to understand different information combinations like range sensor findings and target position.

\section{Conclusion} \label{sec:con}

In this paper, a mapless motion planner was trained end-to-end
through continuous control deep-RL from scratch. We revised the state-of-art continuous deep-RL method so that the training and sample collection can be executed in parallel. By taking the 10-dimensional sparse range findings and the target position relative to the mobile robot coordinate frame as input, the proposed motion planner can be directly applied in unseen real environments without fine-tuning, even though it is only trained in a virtual environment. When compared to the low-dimensional map-based motion planner, our approach proved to be more robust to extremely complicated environments.


\bibliographystyle{IEEEtran}
\bibliography{maplessbib}
\end{document}